\documentclass[letterpaper, 10 pt, conference]{ieeeconf}  

\IEEEoverridecommandlockouts

\usepackage{times}
\usepackage{multicol}
\usepackage[bookmarks=true]{hyperref}
\usepackage{xcolor}
\usepackage{hyperref}
\usepackage{amssymb}
\usepackage{amsmath, amsfonts}
\usepackage{graphicx}
\usepackage{siunitx}
\usepackage{standalone}
\usepackage{booktabs}
\usepackage[ruled,vlined]{algorithm2e}
\usepackage{mdframed}
\usepackage{fancyvrb}
\usepackage{soul}
\usepackage{dsfont,mathabx}
\usepackage[font=footnotesize, skip=5pt]{caption}

\newtheorem{problem}{Problem}

\title{\LARGE \bf
Interactive Human-in-the-loop Coordination of Manipulation Skills\\ Learned from Demonstration
}

\author{Meng Guo$^{1}$ and Mathias B\"urger$^{2}$
\thanks{$^{1}$College of Engineering, Peking University, China. $^{2}$Bosch Center for Artificial Intelligence (BCAI), Germany. Corresponding author: Meng Guo. \texttt{meng.guo@pku.edu.cn}.}}

\begin{document}
  
\maketitle

\maketitle

\begin{abstract}
Learning from demonstration (LfD) provides a fast, intuitive and efficient framework to program robot skills, which has gained growing interest both in research and industrial applications.
Most complex manipulation tasks are long-term and involve a set of skill primitives.
Thus it is crucial to have a reliable coordination scheme that selects the correct sequence of skill primitive and the correct parameters for each skill, under various scenarios.
Instead of relying on a precise simulator, 
this work proposes a human-in-the-loop coordination framework for LfD skills that: 
builds parameterized skill models from kinesthetic demonstrations;
constructs a geometric task network (GTN) on-the-fly from human instructions;
learns a hierarchical control policy incrementally during execution.
This framework can reduce significantly the manual design efforts, while improving the adaptability to new scenes. 
We show on a 7-DoF robotic manipulator that the proposed approach can teach complex industrial tasks such as bin sorting and assembly in less than 30 minutes.
\end{abstract}


\section{Introduction}\label{sec:introduction}
Robots have been making their ways out of the closed fences in industrial factories. 
Collaborative robots (\emph{cobots}) are intended for direct human interactions within a shared space.
This workspace is much more dynamic than the precisely-arranged structures, e.g., assembly lines.
Moreover, the designated tasks are often low in repetition and high in variance.
Namely, the robots should adapt to different tasks and different scenarios. 
For instance, 
the same service robot might be used for cleaning, packing and transportation. 
Two characteristics are essential for such use cases:
(i) the ability to \emph{quickly} programme robot for different tasks;
(ii) the learned task policy should \emph{adapt} automatically to unforeseen situations.
Most motion planners however require precise modeling of the scene and the robot~\cite{lavalle2006planning}, thus difficult to configure for non-technical end-users.
Instead, learning from demonstrations (LfD) provides an intuitive and efficient method to program robot skills even for layman.

\begin{figure}[t!]
    \centering
    \includegraphics[width=0.98\linewidth]{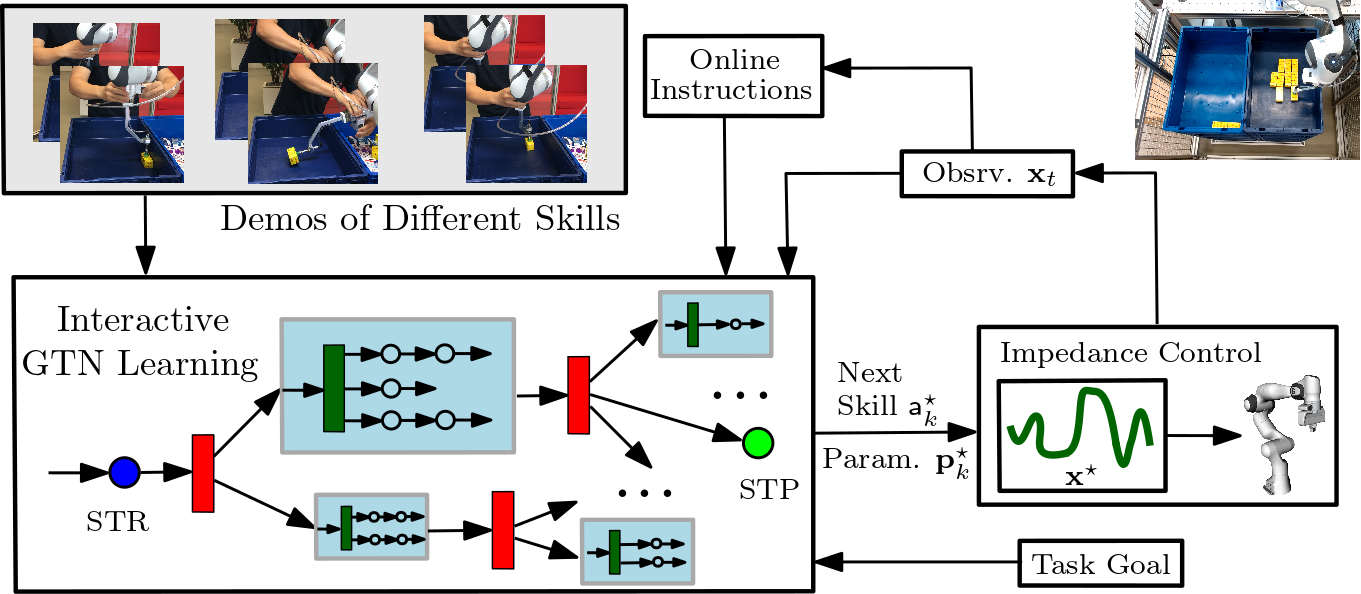}
    \caption{Diagram of the proposed human-in-the-loop skill coordination scheme. 
Once the skill primitives are demonstrated offline, 
the task network and the associated policy are learned online given real-time human instructions and state observations.}
    \label{fig:framework}
    \vspace{-0.15cm}
\end{figure}

Moreover, instead of a single motion, 
complex manipulation tasks often contain multiple branches of skill sequences that share some common skills.
The planning process should generate the right sequence of skills and their parameters under different scenarios. 
For instance, as discussed in later the experiment, 
the bin-picking task involves to pick the object differently from the box, clear it from the corners if needed, re-orient it to reveal the barcode, and show the barcode to the scanner. 
To choose the correct skill sequence is essential for flexible robotic systems across various applications.
Such transitions among the skills and the associated conditions are often difficult and tedious to specify manually.
In fact, both self-adaptation and autonomous decision-making are important design principles of Industry 4.0 systems~\cite{hermann2016design}.

Lastly, besides kinesthetic demonstrations, human operators can also provide \emph{interactive} guidance during online task execution. 
They are often more efficient and effective than offline simulated data,
see~\cite{xin2018accelerating, holzinger2016interactive}.
However, the amount of such inputs should not be excessive and only required under appropriate indications, e.g., when the confidence of a decision process is low. 
In other words, transparency during the interaction is crucial for a human-in-the-loop system.

As shown in Fig.~\ref{fig:framework}, in this work we address these issues by proposing a human-in-the-loop coordination framework for skills learned from demonstrations. 
In particular, the backbone of the coordination framework is a geometric task network (GTN) which consists of the primitive skills as nodes and their transition relations as edges.  
Given a manipulation task, the learned task network can decide not only which skill to execute given the current system state, 
but also the associated parameters. 
The proposed algorithm first constructs the network structure, 
then learns the underlying hierarchical selectors based on the geometric constraints among the robots and the objects.
Human instructions are only required at the first few executions to improve the task network on-the-fly.
Several industrial applications are studied on hardware to validate the performance.

Main contribution of this work is threefold:
(i) a more general task-parameterized model for skills with multiple branches;
(ii) a novel structure as geometric task networks (GTN) for coordinating LfD skills, which is 
fully-explainable regarding the underlying decisions;
(iii) an online, human-in-the-loop and interactive learning algorithm, which is {extremely} data-efficient and intuitive regarding the required human instructions.


\section{Related Work}\label{sec:related}

\subsection{Learning from Demonstration}\label{sec:lfd-review}
Compared with traditional motion planning from~\cite{lavalle2006planning}, learning from demonstration (LfD) is an intuitive way to transfer human skills to robots, see~\cite{calinon2016tutorial, Osa2018Imitation, pathak2018zero}.
Various teaching methods can be used such as kinesthetic teaching in~\cite{calinon2016tutorial}, tele-operation in~\cite{abbeel2004apprenticeship}, and visual demonstration in~\cite{pathak2018zero}. 
Different skill models are proposed to abstract these demonstrations: 
full trajectory of robot end-effector in~\cite{Osa2018Imitation}, 
dynamic movement primitives (DMPs) in~\cite{paraschos2013probabilistic, rozo2016learning},
task-parameterized Gaussian mixture models (TP-GMMs) in~\cite{calinon2016tutorial, Zeestraten17riemannian} which extend GMMs by incorporating observations from different perspectives (so called task parameters),
task-parametrized hidden semi-Markov models (TP-HSMMs) in~\cite{Schwenkel2019Optimizing, rozo2020learning,le2021learning},
and deep neural networks~\cite{pathak2018zero}.
In this work, we adopt the TP-HSMM representation to extract both temporal and spatial features from few human teachings, while allowing generalization over multiple task parameters.

\subsection{Task and Motion Planning}\label{sec:tamp-review}
Task planning focuses on constructing a discrete high-level plan via abstract decision-making (e.g., via logic-reasoning from~\cite{fox2003pddl2}), 
while motion planning addresses the low-level sensing and control problem of a dynamic system, see~\cite{lavalle2006planning}.
The area of task and motion planning (TAMP) attempts to improve the synergies between them.
The planning process over the state graph consists of searching over both discrete skill sequences and the continuous parameters, see~\cite{kaelbling2010hierarchical, konidaris2018skills, Toussaint18diff}.
A conjugate view of the state graph is the so-called skill graph, where instead the nodes are primitive skills and edges are implicit state abstractions, see~\cite{huang2019neural,   hayes2016autonomously,  konidaris2012robot, frazzoli2005maneuver}. 
The work in \cite{hayes2016autonomously} extends the hierarchical task networks (HTN) to conjugate task graph (CTG) without any parameterization on the skill primitives.   
Moreover, \cite{frazzoli2005maneuver} calls such graph as maneuver automaton, which however is \emph{manually} designed instead of learned, whereas~\cite{huang2019neural} require similar structural supervision during training.
The method in~\cite{konidaris2012robot} relies on ``change point'' detection to segment these task demonstrations with simple 2D models,
while~\cite{niekum2015learning} assumes each skill primitive is parameterized to only {one} object frame.
In this work, we adopt this conjugate perspective, but with an embedded hierarchical structure of selectors. 
Moreover, 
both the graph structure and the associated geometric constraints are learned online, without manual specifications. 

\subsection{Human-in-the-loop Learning}\label{sec:hil-review}
Human operators can also provide interactive instructions during online task execution, 
which are often more efficient and effective than offline simulated data,
see~\cite{xin2018accelerating, holzinger2016interactive}.
Human inputs can be of different formats such as qualitative preferences~\cite{holzinger2016interactive}, directly modifying the underlying algorithms~\cite{xin2018accelerating}, additional actions~\cite{mandel2017add}.
In this work, the human operators are only queried when the decision confidence is low, 
and the expected inputs are simple instructions such as the next desired skill or branch name. 

\section{Preliminaries}\label{sec:preliminary}

\subsection{Multi-nomial Classification}\label{subsec:classification}
Multi-nomial or multi-class classification is the problem of classifying instances into one of several classes, see~\cite{bishop2006pattern}.
More precisely, consider the training data $\{(\boldsymbol{y}_m, k_m)\}$ where $\boldsymbol{y}_m\in \mathbb{R}^N$ is the feature vector and $k_m\in \{1,\cdots,K\}$ is the set of possible classes. 
Our goal is to learn a classifier $f:\mathbb{R}^N \rightarrow \{1,\cdots,K\}$ that $f(\boldsymbol{y})$ predicts the most likely class of a new point $\boldsymbol{y}\in \mathbb{R}^N$.
Various algorithms have been proposed such as support vector machines, logistic regression, $k$-nearest neighbors, naive Bayes and neural networks, see~\cite{bishop2006pattern} for details.
An extremely data-efficient yet effective method is to use logistic regression along with the \emph{one-vs-rest} strategy, by maximizing the likelihood of correctly classifying all training data.
More specifically, the probability of $\boldsymbol{y}_m$ belonging to class $k_m$ is given by
\begin{equation}\label{eq:log}
p(k_m \,|\, \boldsymbol{y}_m) = \sigma(\boldsymbol{\beta}^{\intercal} \boldsymbol{y}_m),
\end{equation}
where $\sigma(t)=1/(1+e^{-t})$ is the logistic function for $t\in \mathbb{R}$;
$\boldsymbol{\beta}\in \mathbb{R}^N$ is the model parameter as the weight vector.

Logistic regression is used here (instead of SVMs and DNNs) as: 
(i) it requires much less training data;
(ii) it offers probabilistic measures that can be used as confidence in a decision;
(iii) the decision model is more interpretable as the weighted combination of the feature variables.

\subsection{Skill Model as TP-HSMM}\label{subsec:tp-hsmm}
As proposed in~\cite{calinon2016tutorial, Schwenkel2019Optimizing,rozo2020learning}, 
the skill model~$\boldsymbol{\theta}_{\mathsf{a}}$ of a primitive skill~$\mathsf{a}$ as the TP-HSMM is defined as:
\begin{equation}\label{eq:tp-hsmm}
\boldsymbol{\theta}_{\mathsf{a}} = \left\{ \{a_{hk}\}_{h=1}^K,\, (\mu_k^D, \sigma_k^D),\, \boldsymbol{\gamma}_k \right\}_{k=1}^K,
\end{equation}
where $a_{hk}$ is the transition probability from component~$h$ to component~$k$; 
$(\mu_k^D, \sigma_k^D)$ describe the duration Gaussian model of component~$k$; 
and $\boldsymbol{\gamma}_k$ is component~$k$ of a TP-GMM:
$\boldsymbol{\gamma} = \{\pi_k,\{\boldsymbol{\mu}_k^{p},\boldsymbol{\Sigma}_k^{p}\}_{p=1}^P\}_{k=1}^K$,
where $K$ represents the number of Gaussian components in the mixture model, 
$\pi_k$ is the prior probability of each component, 
and $\{\boldsymbol{\mu}_k^{p},\boldsymbol{\Sigma}_k^{p}\}_{p=1}^P$ are the mean and covariance of the $k$-th Gaussian component within frame~$p$. 
Frames provide observations of the \emph{same} data but from different perspectives, the instantiation of which is called a {task parameter}.
An Expectation-Maximization (EM) method from~\cite{dempster77em} is used to optimize this model.

\section{Problem Description}\label{sec:problem}

Consider a multi-DoF robotic arm within a static and known workspace, of which the end-effector has state $\boldsymbol{r}$ such as its 6-D pose and gripper state.
Also, there are multiple objects of interest denoted by $\mathsf{O}=\{\mathsf{o}_1,\cdots,\mathsf{o}_J\}$. 
Each object is described by its state~$\boldsymbol{p}_{\mathsf{o}}$ such as its 6-D pose.

Moreover, there is a set of \emph{primitive} skills that enable the robot to manipulate these objects, denoted by $\mathsf{A}=\{\mathsf{a}_1,\mathsf{a}_2,\cdots,\mathsf{a}_H\}$.  
For each skill, a human user performs several kinesthetic  demonstrations on the robot.
Particularly, for skill $\mathsf{a}\in \mathsf{A}$, the set of objects involved is given by $\mathsf{O}_{\mathsf{a}} \subseteq \mathsf{O}$ and the set of demonstrations is given by $\boldsymbol{D}_{\mathsf{a}}=\{\mathsf{D}_1,\cdots, \mathsf{D}_{M_{\mathsf{a}}}\}$, where each demonstration $\mathsf{D}_m$ is a {timed} sequence of states $\mathbf{s}\in \mathbf{s}$ that consists of the end-effector state $\boldsymbol{r}$ and object states $\{\boldsymbol{p}_{\mathsf{o}}, \, \mathsf{o}\in \mathsf{O}_{\mathsf{a}}\}$, i.e.,
$\mathsf{D}_m = \big[\mathbf{s}_t\big]_{t=1}^{T_m} = \left[\big(\boldsymbol{r}_t, \,\{\boldsymbol{p}_{t,\mathsf{o}}, \, \mathsf{o}\in \mathsf{O}_{\mathsf{a}}\} \big)\right]_{t=1}^{T_m}$.
Via a combination of these skills, the objects can be manipulated to reach different states. 

We consider a generic manipulation \emph{task}, which however consists of many {instances}. 
Each problem instance is specified by an initial state~$\mathbf{s}_0$ and a set of desired goal states $\mathbf{s}_G$. 
A task is solved when the system state is changed from $\mathbf{s}_0$ to $\mathbf{s}_G$.
Then the problem statement is as follows:
\begin{problem}\label{main-problem}
Given a new task $(\mathbf{s}_0,\,\mathbf{s}_G)$, determine (i) the discrete sequence of skills; 
and (ii) the continuous robot trajectory to execute each skill. \hfill $\blacksquare$
\end{problem}

We are interested in solving complex manipulation tasks where the sequence of desired skills and the associated trajectories change significantly within different scenarios. 

\begin{figure}[t!]
    \centering
    \includegraphics[width=0.45\linewidth]{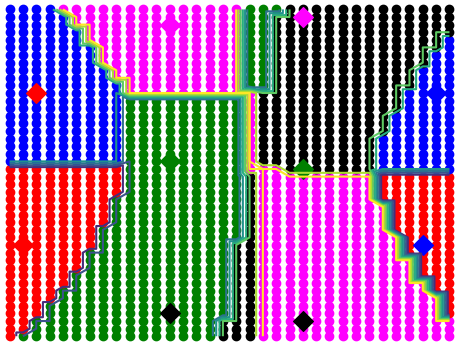}
    \includegraphics[width=0.45\linewidth]{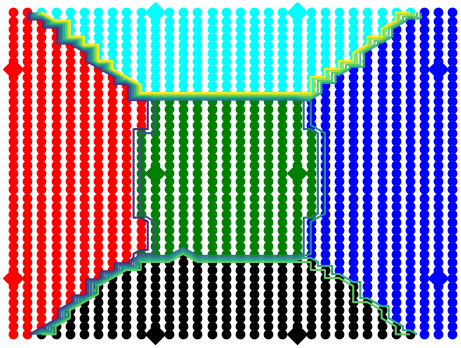}
    \caption{
Comparison between branch selection based on the Gaussian preconditions (left) 
and the proposed selector (right), 
    for the bin-picking skill with 5 branches (four sides and the center). 
    Note that different colors indicate the predicted branches at that sample point (in dots),
 while the \emph{projected} training data are indicated as diamonds (left).}
    \label{fig:branch_compare}
    \vspace{-0.15cm}
\end{figure}

\section{Proposed Solution}\label{sec:solution}
In this section, we present the main components of the proposed solution:
first, we introduce an extension to the current skill model learning;
then, we show how to construct a Geometric Task Network (GTN) for a given task; 
lastly, we describe how both the skill model and the task network can be improved during online execution with human inputs.

\subsection{Primitive Skills Learning}\label{subsec:learn-primitive}

As illustrated in Fig.~\ref{fig:framework}, 
there are often multiple ways of executing the same skill under different scenarios (called \emph{branches}).
For instance, there are five different ways of picking objects from a bin, i.e., approaching with different angles depending on the distances to each boundary.  
Our earlier work~\cite{rozo2020learning} proposed a learning algorithm for TP-HSMM with multiple branches, 
and moreover a precondition model that chooses the best branch based on the first GMMs of all branches.
However, this model requires a large number of demos to cover the area of interest 
and does not generalizes well to new scenarios.
This is mainly due to the usage of Gaussian clustering over few samples in high dimensions. 
Fig.~\ref{fig:branch_compare} shows one example of the bin-picking skill.


\begin{figure}[t!]
    \centering
    \includegraphics[width=0.7\linewidth]{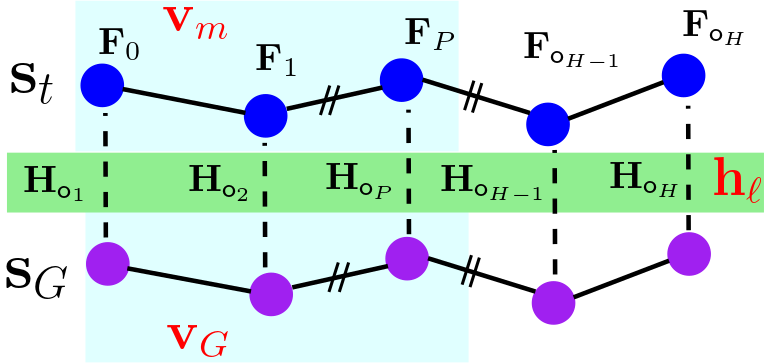}
    \caption{Illustration of the computation of feature vectors $\mathbf{v}_m$ for the edge selector and $\mathbf{h}_\ell$ for the branch selector, given skill frames $\mathbf{F}_p$.}
    \label{fig:frames}
    \vspace{-0.15cm}
\end{figure}

To overcome this limitation, 
we propose a \emph{branch selector} as an extension to the original TP-HSMM model~$\boldsymbol{\theta}_{\mathsf{a}}$ in Sec.~\ref{subsec:tp-hsmm}. 
Consider a skill primitive $\mathsf{a}$ with $M$ demonstrations and $B$ different branches. 
Furthermore, each execution trajectory or demonstration of the skill is denoted by 
$\mathsf{J}_m=\big[\mathbf{s}_t\big]_{t=1}^{T_m},$
which is associated with \emph{exactly} one branch $b_m\in B_{\mathsf{a}}=\{1,\cdots, B\}$. 
Denote by~$\boldsymbol{J}_{\mathsf{a}}$ the set of such trajectories,
initialized to be the set of demos~$\boldsymbol{D}_{\mathsf{a}}$.
As mentioned in Sec.~\ref{subsec:tp-hsmm}, 
the frames associated with~$\mathsf{J}_m$ are computed from the initial state $\mathbf{s}_0$, 
by abstracting the coordinates of the robot and the objects, denoted by:
\begin{equation}\label{eq:frames}
(\mathbf{F}_0,\,\mathbf{F}_1,\cdots,\mathbf{F}_P),
\end{equation}
where $\mathbf{F}_p=(\mathbf{b}^{p},\mathbf{A}^{p})$ is the coordinate of frame $p\in \{1,\cdots,P\}$;
their order can be freely chosen but fixed afterwards.
Then, we can derive the \emph{feature} vector:
\begin{equation}\label{eq:skill-feature}
\mathbf{v}_m = \left( \mathbf{F}_{01},\,\mathbf{F}_{12},\cdots,\mathbf{F}_{(P-1)P}\right),
\end{equation}
where $\mathbf{F}_{ij}=(\mathbf{b}_{ij},\,\boldsymbol{\alpha}_{ij})\in \mathbb{R}^6$ is the relative transformation from frame $\mathbf{F}_i$ to frame $\mathbf{F}_j$:
$\mathbf{b}_{ij}\in \mathbb{R}^3$ is the relative pose and $\boldsymbol{\alpha}_{ij}\in \mathbb{S}^3$ is the relative orientation.
Thus, given $\boldsymbol{J}_{\mathsf{a}}$, we can construct the \emph{training data} for the branch selector:
\begin{equation}\label{eq:branch-data}
\boldsymbol{\tau}_{\mathsf{a}}^{\texttt{B}}=\left\{ (\mathbf{v}_m,\,b_m),\,\forall \mathsf{J}_m\in \boldsymbol{J}_{\mathsf{a}}\right\},
\end{equation}
where $b_m$ is the branch label of trajectory $\mathsf{J}_m$;
$\mathbf{v}_m$ is the associated feature vector.
Then the branch selector, denoted by $\mathcal{C}_{\mathsf{a}}^{\texttt{B}}$, can be learned via any multi-nomial classification algorithms.
As described in Sec.~\ref{subsec:classification}, logistic regression under the ``one-vs-rest'' strategy yields an effective selector from few training samples.
More comparisons are given in Sec.~\ref{sec:experiments}.
Afterwards, given a new scenario with state $\mathbf{s}_t$,
the probability of choosing branch $b$ is given by:
\begin{equation}\label{eq:branch-prob}
\rho_{b} = \mathcal{C}_{\mathsf{a}}^{\texttt{B}}(\mathbf{s}_t,\, b), \;\forall b \in B_{\mathsf{a}},
\end{equation}
where $\rho_{b}\in [0,\,1]$.
Since most skills contain two or three frames, 
the feature vector $\mathbf{v}_m$ in~\eqref{eq:skill-feature} normally has dimension $6$ or $12$.
Fig.~\ref{fig:branch_compare} illustrates much better classification results compared with the Gaussian preconditions~\cite{rozo2020learning}.

\subsection{Task Network Construction}\label{subsec:task-network}

As mentioned in Sec.~\ref{sec:introduction}, complex manipulation tasks often contain various sequences of skills to account for different scenarios. 
A high-level abstraction of such relations is referred as task networks~\cite{hayes2016autonomously}.
A valid plan evolves by transition from one skill to another until the task is solved. 
Even though the graph structure can be sketched easily, 
the \emph{conditions} on these transitions are particularly difficult and tedious to specify manually. 
To overcome this limitation,
we propose a novel coordination structure as geometric task networks (GTNs)~\cite{guo2021geometric}, 
where the conditions are learned from the task execution results.

\begin{figure}[t!]
    \centering
    \includegraphics[width=0.98\linewidth]{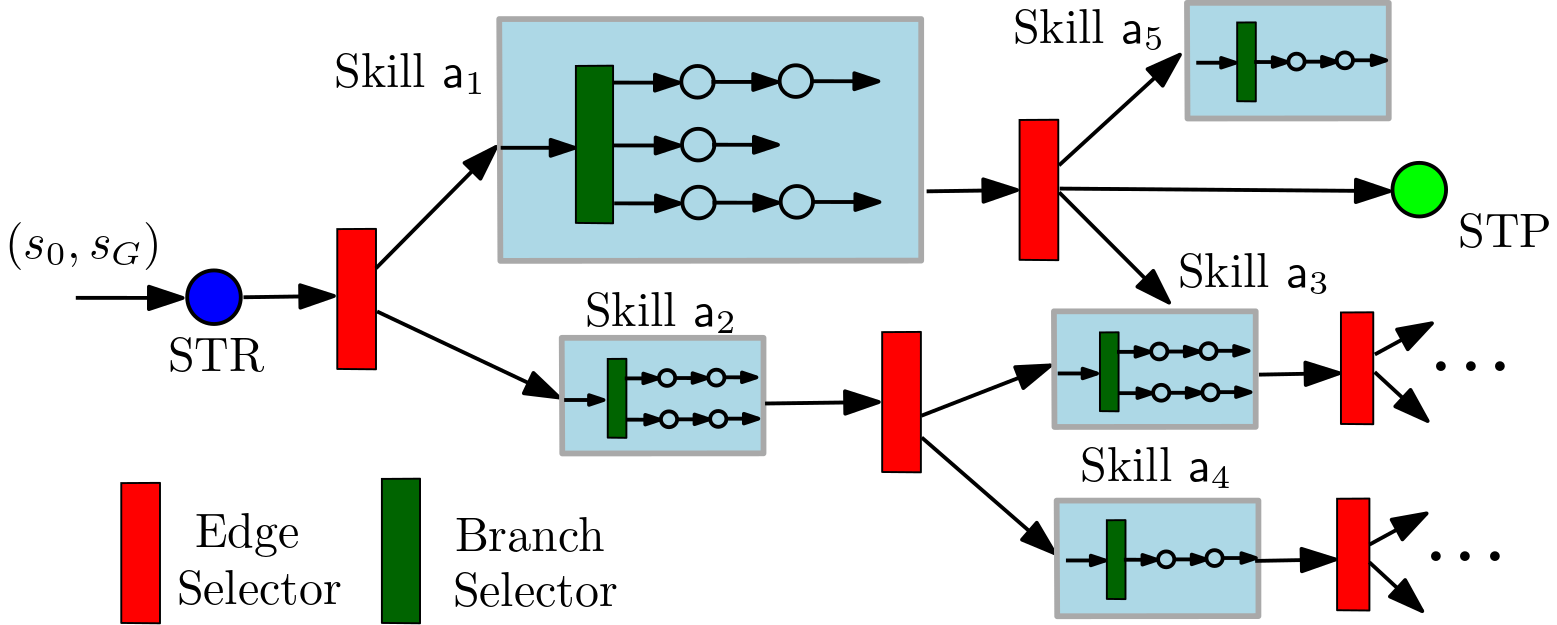}
    \caption{The proposed network structure over primitive skills, 
      where the conditions for selecting edges and branches are critical for the execution.}
    \label{fig:gtn}
    \vspace{-0.15cm}
\end{figure}

\subsubsection{Network Structure}\label{subsubsec:gtn-structure}
As illustrated in Fig.~\ref{fig:gtn}, a GTN has a simple structure defined by the triple $\mathcal{G}=(V,\, E,\, f)$.
The set of nodes~$V$ is a subset of the primitive skills $\mathsf{A}$;
the set of edges $E \subseteq V \times V$ contains the allowed transitions from one skill to another;
the function $f: v \rightarrow \mathcal{C}$ maps each node to a edge selector w.r.t. all of its outgoing edges.
Intuitively, $(V,\, E)$ specifies how skills can be executed consecutively for the given task,
while function $f(v)$ models the different geometric constraints among the objects and the robot,
for the outgoing edges of node $v$.
Note that $f(\cdot)$ is \emph{explicitly} conditioned on both the current system state and the goal state. 
Its detailed model is given in the sequel.


\subsubsection{Learning from Complete Plans}\label{subsubsec:gtn-learn}
Without loss of generality, a complete plan of the considered problem in Problem~\ref{main-problem} is given by the following sequence:
\begin{equation}\label{eq:complete-plan}
\boldsymbol{\xi} =\underline{\mathsf{a}} \mathbf{s}_0 \,\mathsf{a}_0\,\mathbf{s}_1 \,\mathsf{a}_1\, \mathbf{s}_2 \cdots \mathbf{s}_G \overline{\mathsf{a}}, 
\end{equation}
where~$\underline{\mathsf{a}}$ and $\overline{\mathsf{a}}$ are virtual ``start'' and ``stop'' skills, respectively.
For different initial and goal states (i.e., problems) of the same task, the resulting plans can be different. 
Denoted by $\boldsymbol{\Xi}=\{\boldsymbol{\xi}\}$ the set of complete plans for a set of given problems.
Then, for each ``action-state-action'' triple $(\mathsf{a}_n, \mathbf{s}_{n+1}, \mathsf{a}_{n+1})$ within $\boldsymbol{\xi}$,
first, the pair $(\mathsf{a}_{n},  \mathsf{a}_{n+1})$ is added to the edge set $\widehat{E}$ if not already present; 
second, for each \emph{unique} skill transition $(\mathsf{a}_{n},\mathsf{a}_{n+1})$, 
a set of \emph{augmented} states is collected, denoted by~$\widehat{\mathbf{s}}_{\mathsf{a}_{n}\mathsf{a}_{n+1}}=\{\widehat{\mathbf{s}}\}$, 
where $\widehat{\mathbf{s}}= (\mathbf{s}_{n+1},\mathbf{s}_G)$.
Furthermore, for each augmented state $\widehat{\mathbf{s}}_\ell=(\mathbf{s}_t,\mathbf{s}_G)\in \widehat{\mathbf{s}}_{\mathsf{a}_{n}\mathsf{a}_{n+1}}$, 
we derive the following \emph{feature} vector:
\begin{equation}\label{eq:edge-feature}
\mathbf{h}_\ell = (\mathbf{h}_{tG},\, \mathbf{v}_G),\,
\end{equation}
where $\mathbf{h}_{tG}=(\boldsymbol{H}_{\mathsf{r}},\,\boldsymbol{H}_{\mathsf{o}_1},\cdots,\boldsymbol{H}_{\mathsf{o}_H})$, 
where $\boldsymbol{H}_{\mathsf{o}}=(\mathbf{b}_{\mathsf{o}},\,\boldsymbol{\alpha}_{\mathsf{o}})\in \mathbb{R}^6$ is the relative translation and rotation of robot $\mathsf{r}$ and objects $\mathsf{o}_1,\mathsf{o}_2,\cdots, \mathsf{o}_H \in \mathsf{O}_{\mathsf{a}_n}$, 
{from} the current system state $\mathbf{s}_t$ {to} the goal state $\mathbf{s}_G$;
$\mathbf{v}_G$ is the feature vector defined in~\eqref{eq:skill-feature} associated with the goal state $\mathbf{s}_G$.
Note that $\mathbf{h}_\ell$ encapsulates features from both the relative transformation to the goal, and the goal state itself. 
Its dimension is linear to the number of objects relevant to skill $\mathsf{a}_{n}$, as shown in Fig.~\ref{fig:frames}.

Once all plans within $\boldsymbol{\Xi}$ are processed, the GTN $\mathcal{G}$ can be constructed as follows.
First, its nodes and edges are directly derived from $\widehat{E}$.
Then, for each node $\mathsf{a}$, the set of its outgoing edges in $\widehat{E}$ is given by $\widehat{E}_{\mathsf{a}}=\{(\mathsf{a}, \mathsf{a}_\ell) \in \widehat{E}\}$.
Thus the function $f(\mathsf{a})$ returns the edge selector $\mathcal{C}_{\mathsf{a}}^{\texttt{E}}$ over $\widehat{E}_{\mathsf{a}}$.
To compute this selector, we first construct the following training data:
\begin{equation}\label{eq:node-data}
\boldsymbol{\tau}^{\texttt{E}}_{\mathsf{a}} = \left\{ (\mathbf{h}_\ell,\, e),\, \forall \widehat{s}_\ell \in  \widehat{\mathbf{s}}_e,\, \forall e\in \widehat{E}_{\mathsf{a}} \right\},
\end{equation}
where $e$ is the label for an edge $e = (\mathsf{a},\,\mathsf{a}_\ell)\in \widehat{E}_{\mathsf{a}}$;
$\mathbf{h}_\ell$ is the feature vector derived from~\eqref{eq:edge-feature}.
Then the edge selector $\mathcal{C}_{\mathsf{a}}^{\texttt{E}}$ can be learned via the multi-nomial classification algorithms presented in Sec.~\ref{subsec:classification}.
Similar to~\eqref{eq:branch-prob}, 
given a new scenario with state $\mathbf{s}_t$ and the specified goal state $\mathbf{s}_G$,
the probability of choosing edge $e$ is given by:
\begin{equation}\label{eq:edge-prob}
\rho_{e} = \mathcal{C}_{\mathsf{a}}^{\texttt{E}}\left((\mathbf{s}_t,\,\mathbf{s}_G),\, e\right), \;\forall e \in \widehat{E}_{\mathsf{a}},
\end{equation}
where $\rho_{e}\in [0,\,1]$.
Note that $\rho_{e}$ is trivial for skills with only one outgoing edge.

\subsection{Human-in-the-loop Policy Learning and Execution}\label{subsec:learn-policy}
The previous sections present the methods to learn the extended skill model and the task network.
The required training data are execution trajectories of the skill in~\eqref{eq:frames} and complete plans of the task in~\eqref{eq:complete-plan}.
In this section, we show how human instructions during run time can be used to generate such data, 
and thus to learn or improve both the skill model and the task network \emph{on-the-fly}.


\subsubsection{Execute and Update GTN}\label{subsubsec:update-gtn}
The GTN $\mathcal{G}$ is initialized as empty.
Consider a problem instance of the task, namely~$(\mathbf{s}_0, \mathbf{s}_G)$.
The system starts from state $\mathbf{s}_n$ whereas the GTN starts from the virtual start node $\mathsf{a}_n=\underline{\mathsf{a}}$ for $n=0$. 
Then the associated edge selector $\mathcal{C}_{\mathsf{a}_n}^{\texttt{E}}$ is used to compute the probability $\rho_e$ of each outgoing edge $e\in \widehat{E}_{\mathsf{a}_n}$ based on~\eqref{eq:edge-prob}.
Then, the next skill to execute is chosen as:
\begin{equation}\label{eq:next-skill}
\mathsf{a}^\star_{n+1}=
\underset{e\in \widehat{E}_{\mathsf{a}_n}}{\mathrm{argmax}}
\;\left\{\rho_e(\mathbf{s}_n,\,\mathbf{s}_G),\, \text{where}\; \rho_e>\underline{\rho}^{\texttt{E}}\right\},\;
\end{equation}
where $\underline{\rho}^{\texttt{E}}>0$ is a design parameter as the lower confidence bound.
Note that if the current set of outgoing edges is empty, i.e., $\widehat{E}_{\mathsf{a}_n}=\emptyset$, 
{or} the maximal probability of all edges is less than $\underline{\rho}^{\texttt{E}}$, 
the human operator is asked to \emph{input} the preferred next skill $\mathsf{a}^\star_{n+1}$. 
Consequently, an \emph{additional} data point is added to the training data $\boldsymbol{\tau}_{\mathsf{a}_n}^{\texttt{E}}$, i.e., 
\begin{equation}\label{eq:human-input-1}
\boldsymbol{\tau}_{\mathsf{a}_n}^{\texttt{E}} \leftarrow \left( \mathbf{h}(\mathbf{s}_n,\,\mathbf{s}_G),\,(\mathsf{a}_n,\, \mathsf{a}^\star_{n+1}) \right),
\end{equation}
where the feature vector $\mathbf{h}$ is computed based on~\eqref{eq:edge-feature}.
Thus, a new edge $(\mathsf{a}_n,\, \mathsf{a}^\star_{n+1})$ is added to the graph topology $(V, E)$ if not present, 
and the embedded function $f(\cdot)$ is updated by re-learning the edge selectors $\mathcal{C}_{\mathsf{a}_n}^{\texttt{E}}$ given this new data point.

\begin{algorithm}[t]
   \caption{HIL Coordination of LfD Skills} \label{alg:hil-coordination}
   \LinesNumbered
   \DontPrintSemicolon
   \KwIn{$\{\boldsymbol{D}_{\mathsf{a}}, \, \forall \mathsf{a} \in \mathsf{A}\}$, human inputs $\{\mathsf{a}_n^\star, b_n^\star\}$.}
   \KwOut{$\mathcal{G}$, $\{\mathcal{C}^{\texttt{B}}_{\mathsf{a}}\}$, $\boldsymbol{u}^\star$.}
   \tcc{offline learning}
   Learn $\boldsymbol{\theta}_{\mathsf{a}}$ and $\{\mathcal{C}^{\texttt{B}}_{\mathsf{a}}\}$, $\forall \mathsf{a} \in \mathsf{A}$.\;
   Initialize or load existing $\mathcal{G}$.\;
   \tcc{online execution and learning}
   \While{new task $(\mathbf{s}_0, \mathbf{s}_G)$ given}{
       Set $\mathsf{a}_n \leftarrow\underline{\mathsf{a}}$ and $\mathbf{s}_n \leftarrow\mathbf{s}_0$.\;
       \While{$\mathbf{s}_n \neq \mathbf{s}_G$}{
             $\mathcal{G},\, \mathsf{a}_{n+1}=\texttt{ExUpGtn}(\mathcal{G},\, \mathsf{a}_{n},\, (\mathbf{s}_n,\, \mathbf{s}_G),\, \mathsf{a}^\star_{n+1})$.\;
         $ \mathcal{C}^{\texttt{B}}_{\mathsf{a}_{n+1}},\,b_{n+1}=\texttt{ExUpBrs}(\mathcal{C}^{\texttt{B}}_{\mathsf{a}_{n+1}},\, \mathbf{s}_n,\, b^\star_{n+1})$.\;
             Compute $\boldsymbol{u}^\star$ for branch $b_{n+1}$ of skill $\mathsf{a}_{n+1}$.\;
             Obtain new state $\mathbf{s}_{n+1}$. Set $n\leftarrow n+1$.\;
       }
   }
\end{algorithm}
%

\subsubsection{Execute and Update Branch Selector}\label{subsubsec:update-branch}

Now assume that $\mathsf{a}_{n+1}$ is chosen as the next skill. 
Then the branch selector $\mathcal{C}_{\mathsf{a}_{n+1}}^{\texttt{B}}$ is used to predict the probability of each branch $\rho_{b}$, $\forall b\in B_{\mathsf{a}_{n+1}}$
Then, the most likely branch for $\mathsf{a}_{n+1}$ is chosen by:
\begin{equation}\label{eq:next-branch}
b^\star_{n+1}=
\underset{b \in B_{\mathsf{a}_{n+1}}}{\mathrm{argmax}} \;\left\{\rho_{b}(\mathbf{s}_n),\, \text{where} \, \rho_{b} > \underline{\rho}^{\texttt{B}} \right\},
\end{equation}
where $\underline{\rho}^{\texttt{B}}>0$ is another parameter as the lower confidence bound
for the branch selection.
Note that if no branch is found above, then the human operator is asked to input the preferred branch $b^\star_{n+1}$ for skill $\mathsf{a}_{n+1}$.
Consequently, an \emph{additional} data point is added to the training data $\boldsymbol{\tau}_{\mathsf{a}_{n+1}}^{\texttt{B}}$, i.e., 
\begin{equation}\label{eq:human-input-2}
\boldsymbol{\tau}^{\texttt{B}}_{\mathsf{a}_{n+1}} \leftarrow \left( \mathbf{v}(\mathsf{s}_n),\,b^\star_{n+1} \right),
\end{equation}
where the feature vector $\mathbf{v}$ is computed based on~\eqref{eq:skill-feature}.

\subsubsection{Skill Execution}\label{subsubsec:skill-execution}
Once the optimal branch $b^\star$ is chosen for the desired next skill $\mathsf{a}^\star_{n+1}$, 
its trajectory can be retrieved given the skill model~$\boldsymbol{\theta}_{\mathsf{a}_{n+1}}$. 
The retrieval process consists of two steps: First, 
the most-likely sequence of GMM components  within the \emph{desired branch} (denoted by $\boldsymbol{k}^\star_T$) can be computed via the modified Viterbi algorithm proposed in our earlier work~\cite{rozo2020learning}.
Then, a reference trajectory is generated by an optimal controller (e.g., LQG from~\cite{sciavicco2012modelling}) to track this sequence of Gaussians in the task space.
Thus this reference trajectory is then sent to the low-level impedance controller to compute the control signal $\boldsymbol{u}^\star$.
More algorithmic details and extension to Riemannian manifolds are given in~\cite{rozo2020learning}.

Afterwards, the system state is changed to $\mathbf{s}_{n+2}$ with different poses of the robot and the objects, i.e., obtained from  the  state estimation and perception modules. 
Given this new state, the same process is repeated to choose the next skill and its optimal branch, until the goal state is reached.

\subsection{Overall Algorithm}\label{subsubsec:overall-algo}
The overall procedure is summarized in Alg.~\ref{alg:hil-coordination}.
Namely, during the online execution for solving new tasks, 
Sec.~\ref{subsubsec:update-gtn} is followed to execute and update the GTN, 
namely the function $\texttt{ExUpGtn}(\cdot)$ in Line 6, 
with possible human input $\mathsf{a}^\star_n$ if required. 
Once the next skill $\mathsf{a}_{n+1}$ is chosen, 
Sec.~\ref{subsubsec:update-branch} is followed to execute and update the branch selector, 
namely the function $\texttt{ExUpBrs}(\cdot)$ in Line 7, 
with possible human input $b^\star_{n+1}$ if required.
Consequently the GTN and branch selectors are updated and improved via~\eqref{eq:human-input-1} and~\eqref{eq:human-input-2} {on-the-fly}.
Compared with the manual specification of the transition and branching conditions,  
the required human inputs above are more intuitive and easier to specify. 

Lastly, the computation complexity of the logistic regression is $\mathcal{O}(d^2)$, where $d$ is the dimension of the feature vector.
The skill model learning and the LQG control algorithm have polynomial complexity to the robot state dimension. 

\section{Experiments} \label{sec:experiments}
This section presents the experimental validation on a 7-DoF Franka Emika robot arm 
for two different industrial applications: 
first one as a part of an assembly task, and second as a bin-sorting task. 
The arm is extended by a Zivid RGBD sensor for perception 
and a parallel (or suction) gripper. 
Additionally, kinaesthetic teaching can be done directly by guiding the end effector manually. 
The proposed framework is implemented in Python3 under the Robot Operating System (ROS). 
All benchmarks are run on a desktop with an 8-core Intel Xeon CPU. 
Experiment videos can be found in the supplementary file.

\subsection{Workspace and Task Description} \label{subsec:task-description}
\begin{figure}[t!]
    \centering
    \includegraphics[height=0.24\linewidth]{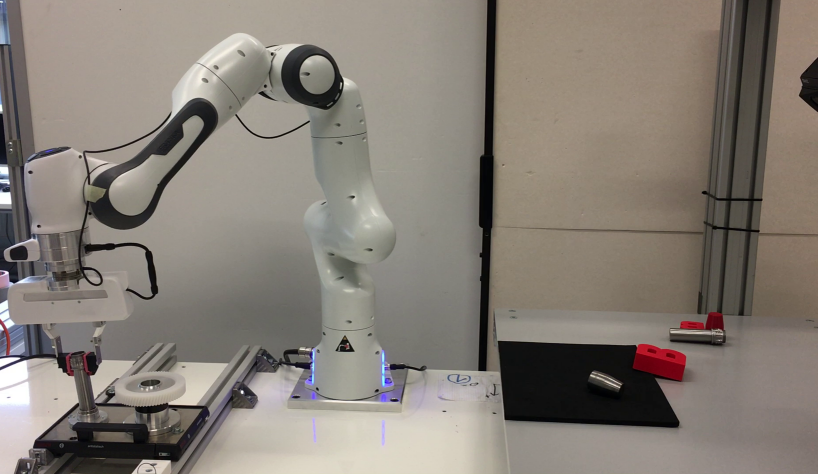}
    \includegraphics[height=0.24\linewidth]{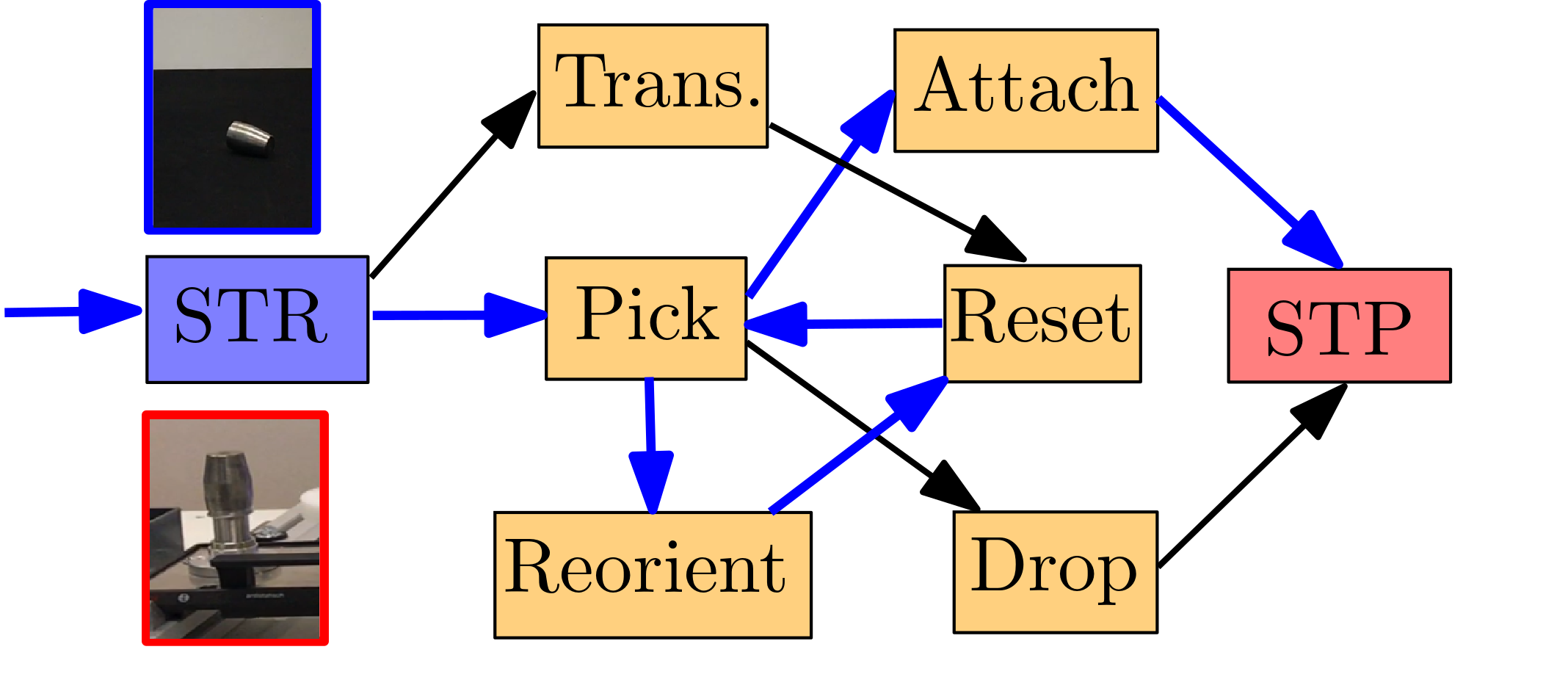}
    \caption{\textbf{Left}: workspace for the assembly task; 
             \textbf{Right}: the learned final GTN with one example plan (in blue).}
    \label{fig:assembly-gtn}
    \vspace{-0.15cm}
\end{figure}

The assembly task was introduced in our previous work~\cite{rozo2020learning}.
For brevity, we omit the detailed description here and refer the readers there.
As shown in Fig.~\ref{fig:assembly-gtn},  a metallic cap is fed onto the inspection platform 
and depending on the result, the robot arm should either attach the cap to the top of a peg 
or drop it into a pallet.
Given different initial states of the cap (lying-flat of standing), 
additional manipulation skills are needed to re-orient and translate the cap before the normal pick and drop skills. 
Note that in our earlier work~\cite{Schwenkel2019Optimizing, rozo2020learning}, 
these sequences are specified \emph{manually} before every execution.

\begin{figure}[t!]
    \centering
    \includegraphics[height=0.24\linewidth]{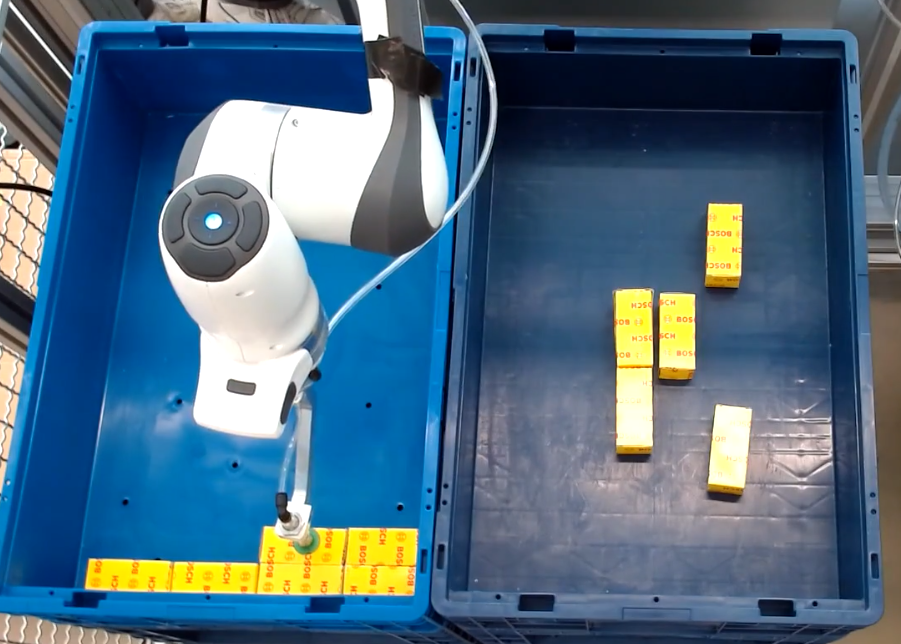}
    \includegraphics[height=0.24\linewidth]{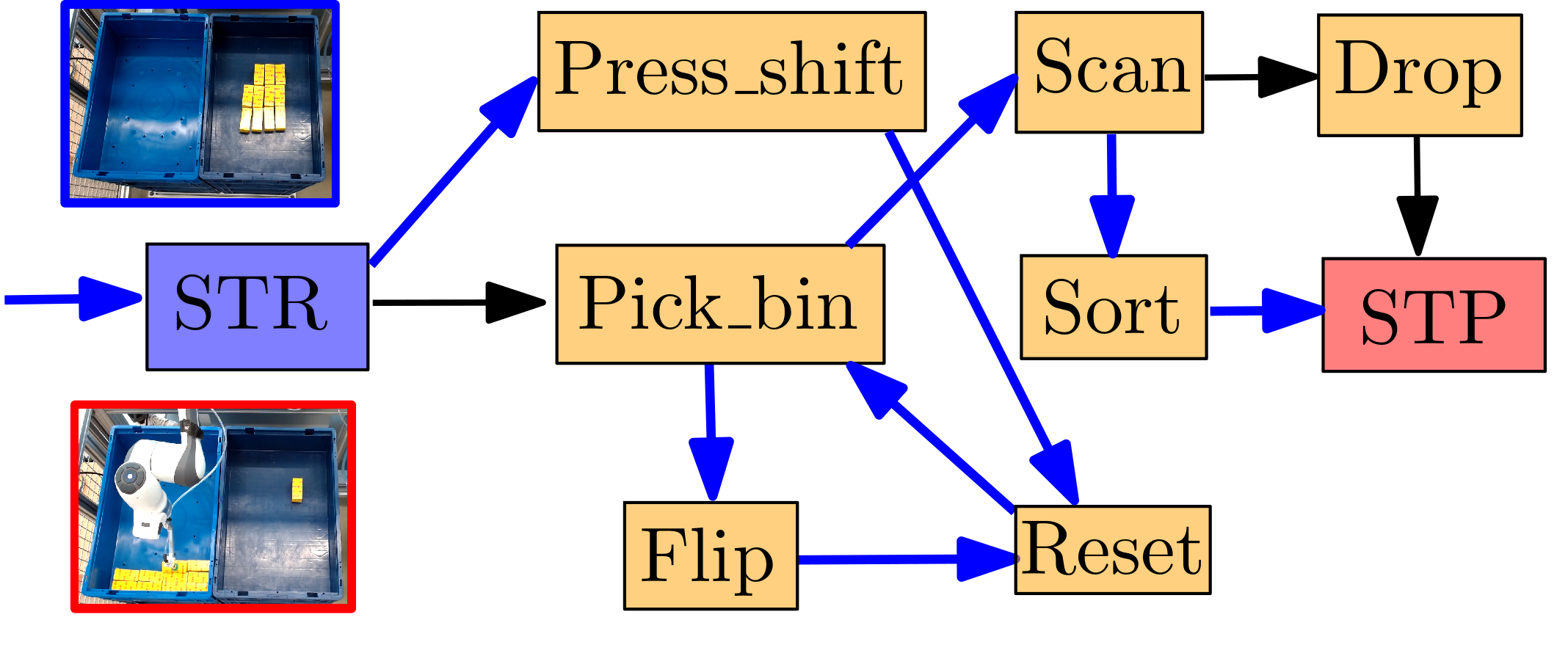}
    \caption{\textbf{Left}: workspace for the bin-sorting task; 
             \textbf{Right}: the learned final GTN with one example plan (in blue).}
    \label{fig:bin-gtn}
    \vspace{-0.15cm}
\end{figure}

Another application is the well-known bin picking and sorting task. 
As shown in Fig.~\ref{fig:bin-gtn}, the goal is to pick unknown objects out of the bin, scan them for product info, and then sort them accordingly. 
Instead of emphasizing performance such as ``pick per hour'', 
we are interested in addressing several corner cases:
(i) the object should be picked with different orientations when close to the bin boundaries;
(ii) the object should be cleared out of the corners before picking;
(iii) the object should be flipped when the barcode overlaps with the suction area;
and (iv) the objects are placed differently given the scanning results.

\subsection{Results} \label{subsec:results}

\subsubsection{Learned Skill Model}\label{subsubsec:skill-model-results}
For the assembly application, in total $5$ primitive skills are taught, i.e., 
\texttt{pick} to pick the cap from the platform with three branches;
\texttt{re\_orient} to re-orient the cap from lying flat to standing with two branches;
\texttt{translate} to translate the cap to the platform boundary while standing;
\texttt{attach} to attach the cap to the peg;
\texttt{drop} to drop the cap with two branches.
Similarly, for the bin sorting application, in total $6$ primitive skills are taught, i.e.,
\texttt{pick\_bin} to pick any object from bin with five branches. 
\texttt{scan} to scan the object; 
\texttt{press\_shift} to press and shift any object out of the corners with four branches;
\texttt{flip} to flip the object;
\texttt{drop\_bin} to drop the picked object into another bin;
\texttt{sort} to arrange the picked object in rows.

Due to the new branch selector proposed in Sec.~\ref{subsec:learn-primitive}, 
the number of demos needed is much reduced compared with our previous work~\cite{Schwenkel2019Optimizing, rozo2020learning}.
In average $6$ demos are performed for each skill and the associated skill model with the branch selector is learned in $0.2s$.
Afterwards, the model of each skill is verified independently under different scenes.
For instance, the \texttt{pick} skill are demonstrated $2$ times for each branch,
and the feature vector $\mathbf{v}$ has dimension $7$.

\begin{figure}[t!]
    \centering
    \includegraphics[width=0.98\linewidth]{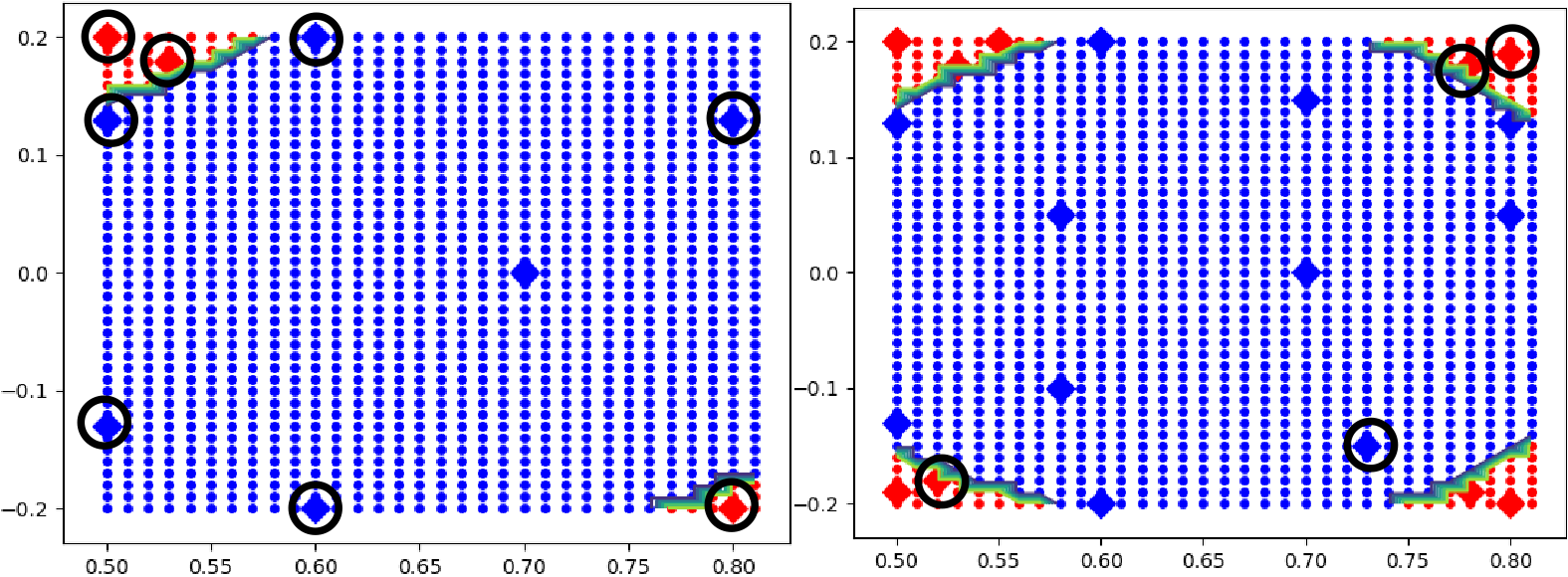}
    \caption{Evolution of the edge selector for node $\texttt{STR}$ in the GTN for bin-sorting in Fig.~\ref{fig:bin-gtn}.
    There are two modes projected onto the $x-y$ plane: 
    \texttt{pick\_bin} skill (in red) and \texttt{press\_shift} skill (in blue). 
    Human instructions are requested at the samples marked in circles, 
    while autonomous predictions are marked in diamonds.}
    \label{fig:change-edge-selector}
    \vspace{-0.15cm}
\end{figure}

\subsubsection{Learned Task Network}\label{subsubsec:task-network-results}
The proposed GTN is learned by following Alg.~\ref{alg:hil-coordination} for each application.
Various problem instances of $(\boldsymbol{s}_0, \boldsymbol{s}_G)$ are defined in the learning process.
Human instructions are requested for the next desired  skill and branch.
For instance, as shown in Fig.~\ref{fig:assembly-gtn},
if the cap is standing and the goal is to drop it into the pallet, 
the skill sequence is \texttt{pick}, \texttt{translate}, \texttt{pick} again, and finally \texttt{drop}.
As shown in Fig.~\ref{fig:bin-gtn}, 
if the object is close to one corner with the barcode on the top, 
the sequence is \texttt{press\_shift},  \texttt{pick\_bin}, \texttt{flip}, \texttt{scan}, and  \texttt{drop} into another bin.

\begin{figure}[t!]
    \centering
    \includegraphics[width=0.98\linewidth]{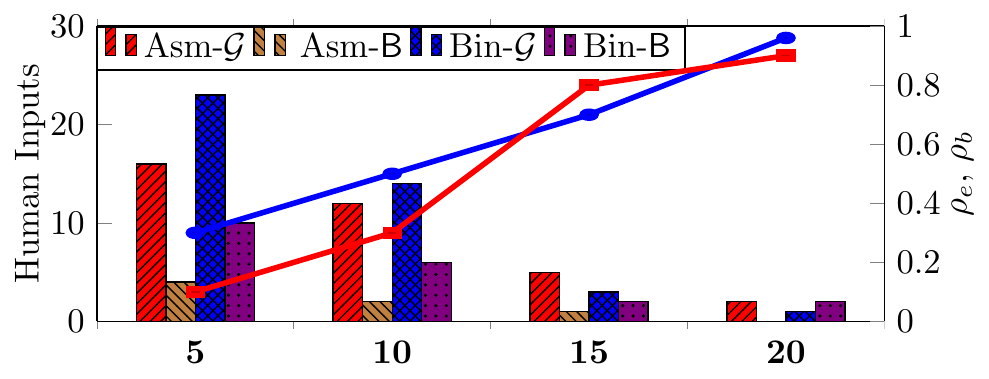}
    \caption{$x$-axis: number of task executions. Left-$y$-axis: number of human instructions requested for GTN ($\mathcal{G}$) and branch selector ($\mathsf{B}$), for assembly task \texttt{Asm} and bin-sorting task \texttt{Bin}.
    Right-$y$-axis: evolution of the lowest confidence in the GTN execution for both tasks.}
    \label{fig:human-input-change}
    \vspace{-0.15cm}
\end{figure}

Each time a human instruction is given, either the GTN or the branch selector is updated via Line~$6-7$ in Alg.~\ref{alg:hil-coordination}.
Fig.~\ref{fig:change-edge-selector} shows an example how the edge selector of node GTN evolves within the execution of the first five executions.
Moreover, Fig.~\ref{fig:human-input-change} records how many human inputs are required for the GTN $\mathcal{G}$, and the branch selectors $\{\mathcal{C}^{\texttt{B}}\}$ are updated during the whole run.
Notice that the topology of $\mathcal{G}$ is quickly learned while the edge and branch selectors are improved whenever a new scene is experienced with a low confidence score. 
Fig.~\ref{fig:human-input-change} also shows how the lowest confidences for selecting edges and branches increase with time as the GTN is improved, 
where both lower bounds $\underline{\rho}^\texttt{B}$ and $\underline{\rho}^\texttt{E}$ are set to $0.8$.
In the end, the success rate is close to $100\%$ with full autonomy for both applications, where most failures are caused by execution and perception errors. 
The final learned GTNs for both are shown in Fig.~\ref{fig:assembly-gtn} and~\ref{fig:bin-gtn}.

\subsection{Comparison} \label{subsec:comparison}
The proposed scheme (\textbf{GTN} for short) is compared to the following baselines:
(i) the vanilla TP-HSMM scheme as stated in~\cite{calinon2016tutorial, rozo2020learning} (\textbf{TPH} for short), i.e., in combination with proposed GTN but without the proposed branch selector;
(ii) the \emph{full} system state (\textbf{FUL} for short) is used as the feature vector for the branch selector in~\eqref{eq:skill-feature} and the edge selector in~\eqref{eq:edge-feature}, 
i.e., instead of the relative frames;
(iii) a task and motion planner (\textbf{TMP} for short) that searches in simulation over the system state space for each new task $(\boldsymbol{s}_0, \boldsymbol{s}_G)$;
(iv) a completely manual design of the branch and edge selection (\textbf{MAN} for short), 
i.e., by specifying the rules for each case.

As summarized in Table~\ref{table:compare}, 
\textbf{TMP} does not learn from past solutions and requires the longest solution time.
For certain problems where the sequence has more than $4$ skills, it can not solve them in reasonable time ($10$ min).
Moreover, \textbf{TPH} performs relatively well for predicting the skill sequence, however fails at executing the skill due to choosing the wrong branch, especially when the scenes are different from the demos.
Notably, \textbf{FUL} learns not only slower but also performs worse in predicting both the edges and the branches, compared with \textbf{GTN}.
The plausible explanation is that the \emph{relative transformation} in~\eqref{eq:skill-feature} and~\eqref{eq:edge-feature} is difficult to capture with linear or even nonlinear kernels such as RBF~\cite{bishop2006pattern}.
Last but not least, the manual rules in \textbf{MAN} are much harder to design than the string inputs required by our scheme. 
Particularly, for both applications, boundaries on orientations of \emph{different} objects are 
often transformed to Euler angles, which however are often ill-posed;
In addition, such rules are hard to cover the complete state space and thus some corner cases are not defined. 
\begin{table}[t]
\begin{center}
\begin{adjustbox}{height=0.15\linewidth}
\begin{tabular}{ccccc}
\toprule
Methods & L-time[$s$] & S-time[$s$] & R-skill/task & H-design[min]\\
\midrule
\textbf{GTN} & 1.5 & 0.2 & 0.95/0.9 & 2 \\
{TPH} & 0.1 & 0.8 & 0.8/0.6 & N/A \\
{FUL} & 10 & 0.5 & 0.3/0.2 & 10 \\
{TMP} & N/A & $>300$ & 0.6/0.5 & N/A \\
{MAN} & N/A & 0.1 & 0.9/0.9 & 20 \\
\bottomrule
\end{tabular}
\end{adjustbox}
\caption{Comparison: learning time, solution time, success rate for each skill and the complete task, and the total time to design human inputs, of all baselines for the assembly task.}
\label{table:compare}
\end{center}
\vspace{-0.35cm}
\end{table}

\section{Conclusion} \label{sec:conclusion}

This work proposes a human-in-the-loop coordination framework for LfD skills that
constructs and learns a geometric task network on-the-fly from human instructions.
The resulting framework is data-efficient and intuitive even for non-technical operators.  
Future work involves the composition of various GTNs and interactive teaching.



\bibliographystyle{IEEEtran}
\bibliography{contents/references}

\end{document}